\documentclass[10pt,journal,compsoc]{IEEEtran}
\usepackage{cite}
\usepackage{amsmath,amssymb,amsfonts}
\usepackage{algorithm,algorithmic}
\usepackage{graphicx}
\usepackage{textcomp}
\usepackage{comment}
\usepackage{threeparttable}
\usepackage{hyperref}
\usepackage{threeparttable}
\usepackage{multirow}
\usepackage{booktabs}
\usepackage{subfig}

\def\BibTeX{{\rm B\kern-.05em{\sc i\kern-.025em b}\kern-.08em
    T\kern-.1667em\lower.7ex\hbox{E}\kern-.125emX}}
    
\begin{document}
\title{Otago Exercises Monitoring for Older Adults by a Single IMU and Hierarchical Machine Learning Models}
\author{Meng~Shang,
        Lenore~Dedeyne,
        Jolan~Dupont,
		Laura~Vercauteren,
		Nadjia~Amini,
		Laurence~Lapauw,
		Evelien~Gielen,
		Sabine~Verschueren,
		Carolina~Varon,
		Walter~De Raedt,
        and~Bart~Vanrumste% <-this % stops a space
\thanks{M. Shang, C. Varon and B. Vanrumste are with KU Leuven, STADIUS, Department of Electrical Engineering, 3000 Leuven, Belgium, e-mail: meng.shang@kuleuven.be.}% <-this % stops a space
\thanks{M. Shang and W. De Raedt are with Imec, Kapeldreef 75, 3001 Leuven, Belgium.}% <-this % stops a space
\thanks{M. Shang and B. Vanrumste are with KU Leuven, e-Media Research lab.}
\thanks{L. Dedeyne, J. Dupont, L. Vercauteren, N. Amini, L. Lapauw, E. Gielen are with Geriatrics \& Gerontology, Department of Public Health and Primary Care, KU Leuven, Belgium.}
\thanks{J. Dupont and E. Gielen are with Department of Geriatric Medicine, UZ Leuven, Belgium.}
\thanks{S. Verschueren is with the Musculoskeletal Rehabilitation Research Group, Department of Rehabilitation Sciences, KU Leuven.}
}

\maketitle

\begin{abstract}
Otago Exercise Program (OEP) is a rehabilitation program for older adults to improve frailty, sarcopenia, and balance. Accurate monitoring of patient involvement in OEP is challenging, as self-reports (diaries) are often unreliable. The development of wearable sensors and their use in Human Activity Recognition (HAR) systems has lead to a revolution in healthcare. However, the use of such HAR systems for OEP still shows limited performance. The objective of this study is to build an unobtrusive and accurate system to monitor OEP for older adults. Data was collected from 18 older adults wearing a single waist-mounted Inertial Measurement Unit (IMU). Two datasets were recorded, one in a laboratory setting, and one at the homes of the patients. A hierarchical system is proposed with two stages: 1) using a deep learning model to recognize whether the patients are performing OEP or activities of daily life (ADLs) using a 10-minute sliding window; 2) based on stage 1, using a 6-second sliding window to recognize the OEP sub-classes. Results showed that in stage 1, OEP could be recognized with window-wise f1-scores over 0.95 and Intersection-over-Union (IoU) f1-scores over 0.85 for both datasets. In stage 2, for the home scenario, four activities could be recognized with f1-scores over 0.8:  \textit{ankle plantarflexors}, \textit{abdominal muscles}, \textit{knee bends}, and \textit{sit-to-stand}. These results showed the potential of monitoring the compliance of OEP using a single IMU in daily life. Also, some OEP sub-classes are possible to be recognized for further analysis.
\end{abstract}


\section{Introduction}
\label{sec:introduction}

\IEEEPARstart{O}{lder} adults suffer from higher fall risk and consequent fall-related injuries. Annually, 24\% to 40\% of community-dwelling older persons fall, of which 21\% to 45\% fall regularly \cite{milat2011prevalence}. Many of these persons need to spend a long time recovering from injuries. Older adults with certain diseases such as sarcopenia and obesity could suffer from even higher fall risk \cite{landi_sarcopenia_2012}. With the increase of the older population, this problem increases the costs of healthcare systems \cite{noauthor_world_nodate}.\par

To reduce fall risk, the Otago Exercise Program (OEP) was developed for older adults. OEP contains a series of balance, strength, and walking exercises, and it has been proven to reduce fall risk and mortality for community-dwelling older adults \cite{thomas_does_2010}. There are more than twenty types of exercises (sub-classes) in the OEP and the participants need to perform these sub-classes sequentially. Older adults participating in the OEP are requested to perform the exercises twice or three times a week within some consecutive weeks and their compliance with performing OEP is monitored by means of self-reports (diaries) \cite{mat_effect_2018,almarzouki_improved_2020}. However, this is not an accurate method to assess their involvement in the OEP, since they sometimes inaccurately remember or report information about their past experiences. Besides, the diaries only record the duration of performing the OEP rather than which OEP sub-classes are performed. \par

An alternative to the self-reports is to apply wearable sensors combined with machine learning techniques. Although Internet-of-Things (IoT) sensors have been widely applied for Human Activity Recognition (HAR), their use for recognizing the OEP has rarely been investigated. In fact, to date, only two studies have used wearable sensors for this purpose. The first study \cite{bevilacqua_human_2019} applied five sensors to recognize the OEP, which were not user-friendly for older adults to wear in daily life. Although the f1-scores for most activities were high, the recruited subjects only included healthy young adults. Besides, the study did not try to distinguish between the OEP and Activities of Daily Life (ADLs). The second study \cite{dedeyne_exploring_2021} applied one Inertial Measurement Unit (IMU) on the waist to collect data from older adults, and two problems were investigated: 1) to distinguish between OEP and ADLs, and 2) to recognize OEP sub-classes. However, for both tasks, the f1-scores did not exceed 0.8.\par

Considering the gaps in the previous studies, this study investigates OEP recognition for community-dwelling older adults using a single IMU on the waist. The aim is to build an offline HAR system for the therapists to analyze their long-term compliance with OEP. The OEP was recognized in two stages: 1) using a large sliding window to recognize the occurrence of OEP and ADLs and 2) using a small sliding window to recognize OEP sub-classes. The data were collected in both a lab scenario and home scenarios from older adults to validate the proposed system. Machine learning and deep learning models were applied and compared in the two stages to optimize the recognition performance.\par

The contributions of this study are:

\begin{enumerate}
  \item For both lab and home scenarios, a single wearable IMU was applied to recognize:
  \begin{enumerate}
  	\item the OEP from ADLs
  	\item some sub-classes of OEP
  \end{enumerate}
  To date, this is the least obtrusive wearable system that could be applied in daily life.
  \item A hierarchical architecture was designed for activity classification, where a large sliding window was used to recognize OEP while a following small sliding window was used to recognize OEP sub-classes.
  \item A state-of-the-art deep learning model combining two different neural network architectures was built to classify OEP and ADLs with the results outperforming other models.
  \item A new approach for defining OEP sub-classes. Initially, level 1 OEP sub-classes were labeled and classified. These sub-classes were further classified into level 2 sub-classes. This approach led to a reduction in the number of classes used in machine learning models, resulting in improved classification performance.
\end{enumerate}

The paper is organized as follows: Section~\ref{sec:Related} reviews the state of the art in the field of HAR. Section~\ref{sec:method} introduces the datasets and the implementation of the proposed system. Section~\ref{sec:results} presents the experimental results and section~\ref{sec:discussion} discusses the results. Finally, section~\ref{sec:conclusion} concludes the paper and proposes future work.

\section{Related works}
\label{sec:Related}

In this section, two topics are discussed: the sliding window techniques and machine learning models that have been used in HAR systems.

\subsection{sliding window}
A common pre-processing method for HAR is to segment the signals into smaller pieces with the same length for feature extraction \cite{wang_survey_2019}. This method is called the sliding window technique. The size of the applied sliding window is important for classification results and dependent on the characteristics of the activities to be recognized. For example, small sizes such as 2s could be applied to recognize static and periodic activities (e.g. walking, sitting, etc.) \cite{li_human_2022}. For these activities, it is suggested that the window size should cover at least one cycle of the activities \cite{wang_survey_2019}. On the other hand, a sliding window size of as large as 1 minute could be applied for activities involving more gestures (e.g. eating) \cite{ellis_multi-sensor_2014}. Since these activities include many sub-activities, it is harder to define the appropriate window size. Sometimes the sliding window is applied with an overlap rate to offer more segments.

\subsection{Machine Learning Models}
Traditionally, feature-based machine learning has been used in HAR systems. For this technique, hand-crafted features were important to be extracted from the raw signals. Typical hand-crafted features include time-domain and frequency-domain features \cite{wang_survey_2019}. These features might not capture all relevant information for classification.\par

Thanks to the evolution of computational power, deep learning models have been applied for the representation of raw signals. With the features extracted from the deep learning models, the recognition performance increased in many cases compared with the models based on hand-crafted features \cite{ferrari_hand-crafted_2019, shakya_comparative_2018}. However, deep learning models normally need more data for training. \par

Convolutional neural networks (CNN) could extract neighboring information from adjacent samples by convolutional layers to be classified by the fully-connected layers. Recently, such architecture is popular for sensor signals classification/regression by performing 1-D \cite{lee_human_2017} or 2-D \cite{wagner_activity_2017} convolution. Recurrent neural networks (RNN), on the other hand, extract temporal features from the time series. For HAR systems, Long Short-Term Memory (LSTM) as a type of RNN units has also been proven efficient for long time series \cite{zhao_deep_2018}. Besides, Transformers have also been explored for HAR using attention mechanism \cite{dirgova2022wearable,trujillo2023accuracy}. The mechanism searches for correlation between features by mapping a query and a set of keys, which makes it efficient for long-time series recognition.\par

With the development of CNN and LSTM, a new architecture called CNN-LSTM was developed for HAR systems \cite{mutegeki_cnn-lstm_2020, mekruksavanich_smartwatch-based_2020}. The features extracted by the convolutional layers were applied as the input for the LSTM units. Then the LSTM units further extract the temporal features for classification. This architecture has been validated for many datasets with better recognition performance than both CNN and LSTM \cite{mutegeki_cnn-lstm_2020, mekruksavanich_smartwatch-based_2020}. It also outperformed Transformers in recent research for HAR \cite{trujillo2023accuracy}. Based on this architecture and Bi-directional-LSTM (BiLSTM) layers, the CNN-BiLSTM was further developed \cite{challa_multibranch_2022, hoai_thu_hihar_2021}. This architecture was applied to learn from both forward and backward time series. \par

In this work, a deep learning architecture was applied to distinguish OEP and daily activities.

\section{Methodology}
\label{sec:method}

\subsection{Data Collection}
This study received approval from the Ethics Committee Research UZ/KU Leuven (S59660 and S60763). Written informed consent was obtained from all participants prior to study participation.\par

In the experiments, community-dwelling older adults (aged 65 and older) performed modified OEP while wearing a McRoberts MoveMonitor+ (McRoberts B.V., Netherlands) with a 9-axis IMU inside. To make a user-friendly system, the subjects were asked to wear the device loosely and comfortably on the waist, with possible misplacement, and without any calibration. The OEP is a rehabilitation program designed to reduce fall risk in older adults. The original program includes multiple sub-classes that need to be performed sequentially. The sub-classes are shown in Table~\ref{tab:OEPlabels}. Besides, subjects also performed other ADLs. Two datasets were collected in the study in different scenarios:

\begin{table}[!t]
\scriptsize
\caption{The labels from level 1 OEP (six classes), level 2 OEP (15 classes), and original OEP}
\label{tab:OEPlabels}
\centering
\begin{tabular}{|l|l|p{2.8cm}|}
\hline
\textbf{Level 1   OEP}            & \textbf{Level 2 OEP} & \textbf{Original OEP}                                                                              \\ \hline
\multirow{6}{*}{Genral walking}   & Marching             & Marching                                                                                           \\ \cline{2-3} 
                                  & Backwards walking    & Backwards walking, Heel toe backwards walking                                                      \\ \cline{2-3} 
                                  & Forwards walking     & Tandem walking, Toe walking                                                                        \\ \cline{2-3} 
                                  & Walking and turn     & Walking and turn                                                                                   \\ \cline{2-3} 
                                  & Sideways walking     & Sideways walking                                                                                   \\ \cline{2-3} 
                                  & Stairs walking       & Stairs walking                                                                                     \\ \hline
\multirow{5}{*}{General standing} & Back moblizer        & Back moblizer                                                                                      \\ \cline{2-3} 
                                  & Ankle plantarflexors & Ankle plantarflexors                                                                               \\ \cline{2-3} 
                                  & Ankle dorsiflexors   & Ankle dorsiflexors                                                                                 \\ \cline{2-3} 
                                  & Knee bends           & Knee bends                                                                                         \\ \cline{2-3} 
                                  & Static standing      & Head mobilizer, Neck mobilizer, Hip abductor, Knee flexors, Tandem stance, One leg stance          \\ \hline
Trunk moblizer                    & Trunk moblizer       & Trunk moblizer                                                                                     \\ \hline
Abdominal muscles                 & Abdominal muscles    & Abdominal muscles                                                                                  \\ \hline
Sit to stand                      & Sit to stand         & Sit to stand                                                                                       \\ \hline
Sitting                           & Sitting              & Head moblizer, Neck moblizer, Ankle moblizer, Knee extensor, Plantar and knee flexor, Hip extensor \\ \hline
\end{tabular}
\end{table}

\subsubsection{Dataset 1: Lab}
The dataset was collected in the lab. The subjects performed the OEP and ADLs with instructions from the researchers certified as OEP leaders. The data of OEP and ADLs were collected either on two separate days or consecutively on one day. The ADLs included walking, walking stairs, sitting, standing, and indoor cycling. 
\subsubsection{Dataset 2: Home}
This dataset was collected at home with videos recorded. With the camera turned on, the subjects wore the device by themselves and followed a booklet with instructions to perform OEP. Before and/or after the OEP, the subjects also randomly performed ADLs at home, out of the camera view. Therefore, the ADLs were not observed but still labeled while the subjects were wearing the IMU.

The recruited older adults were (pre-)sarcopenic or non-sarcopenic (defined by EWGSOP1 \cite{cruz2010sarcopenia}). The detailed information is shown in Table~\ref{tab:subinfo}. The recruited subjects of the two datasets were different. For both datasets, each OEP sub-class was recorded and annotated based on videos. Between the sub-classes, the subjects were not instructed nor monitored. They could be practicing the exercises, sitting for a rest, or walking out of the camera view.

\begin{table}[]
\caption{Subjects information from the two datasets}
\label{tab:subinfo}
\centering
\begin{tabular}{|l|l|l|l|}
\hline
          &  number & age             & gender \& sarcopenia status                                                                                   \\ \hline
dataset 1 & 11              & 84.09$\pm$5.28 & \begin{tabular}[c]{@{}l@{}}5 females (2   (pre-)sarcopenia) \\ 6 males (3 (pre-)sarcopenia)\end{tabular} \\ \hline
dataset 2 & 7               & 69.43$\pm$2.92 & \begin{tabular}[c]{@{}l@{}}4 females (2 (pre-)sarcopenia) \\ 3 males   (2 (pre-)sarcopenia)\end{tabular}  \\ \hline
\end{tabular}
\end{table}

\subsection{OEP exercises annotation}
\label{subsec:OEPlabel}

\subsubsection{level 1 OEP}
Directly classifying the OEP sub-classes was difficult for the machine learning models, due to the relatively small number of training examples and large number of classes. Therefore, the OEP sub-classes were merged according to the characteristics of the exercises, in order to build a hierarchical system, as proposed in \cite{kondo2021sensor}. This technique reduced the number of classes while maintaining the same number of training examples. As shown in Table~\ref{tab:OEPlabels}, some sub-classes were merged as \textit{general walking}, and some were merged as \textit{general standing}. To distinguish from \textit{static standing}, \textit{general standing} also included the exercises that required irregular trunk movement while standing (e.g. \textit{ankle plantarflexors}). As a result, there were six level 1 OEP exercises. 

\subsubsection{level 2 OEP}
Although there were 26 original OEP sub-classes as shown in table~\ref{tab:OEPlabels} \cite{dedeyne_exploring_2021}, a single IMU on the waist could not distinguish all of them. Since many OEP sub-classes do not involve trunk movements (e.g. head mobilizer, neck mobilizer), they were merged as \textit{sitting} and \textit{static standing}. After merging, there were 15 OEP sub-classes to be recognized as shown in Table~\ref{tab:OEPlabels}. These sub-classes were called level 2 OEP (top level). The recognition of level 1 OEP and level 2 OEP were cascaded to improve performance. The details will be introduced in section~\ref{subsec:systemoverview}.

\subsubsection{Transition activities}
During the OEP, the subjects followed the instructions either from the researchers (dataset 1) or from the booklets (dataset 2). Before each OEP sub-class, they could be practicing the exercises, sitting for a rest, or walking out of the camera view. These activities could not be properly monitored or labeled. Therefore, the activities between each two OEP sub-classes were annotated as transition activities. Different from ADLs that happened outside the whole OEP, transition activities were short activities happening within the OEP program. These activities were not included in the training set. \par

In these annotation steps, if a segment contained multiple activities, it was labeled as the majority class and removed when training the models. The duration and number of segments (by a 6-second window with 50\% overlap) are shown in Table~\ref{tab:duration}. The duration of transition activities was observed longer than OEP in dataset 1, which can be attributed to the older age and increased need for rest among the subjects. For dataset 2, \textit{walking stairs} was not performed because of the limitation of camera installation.

\begin{table}[!t]
\caption{The total duration (minutes) and number of segments (by a 6-second window with 50\% overlap) in the datasets}
\center
\label{tab:duration}
\begin{tabular}{|l|ll|ll|}
\hline
        & \multicolumn{2}{l|}{duration}    & \multicolumn{2}{l|}{number of  segments}    \\ \hline
        & \multicolumn{1}{l|}{dataset 1} & dataset 2 & \multicolumn{1}{l|}{dataset 1} & dataset 2 \\ \hline
OEP     & \multicolumn{1}{l|}{196.91}    & 219.29    & \multicolumn{1}{l|}{3436}      & 1582      \\ \hline
transition & \multicolumn{1}{l|}{543.94}    & 89.46     & \multicolumn{1}{l|}{10472}     & 4102      \\ \hline
ADLs      & \multicolumn{1}{l|}{947.92}    & 828.7     & \multicolumn{1}{l|}{18958}     & 23612     \\ \hline
\end{tabular}
\end{table}

\subsection{System overview}
\label{subsec:systemoverview}

\begin{figure}[!t]
\centering
\includegraphics[width=0.7\columnwidth]{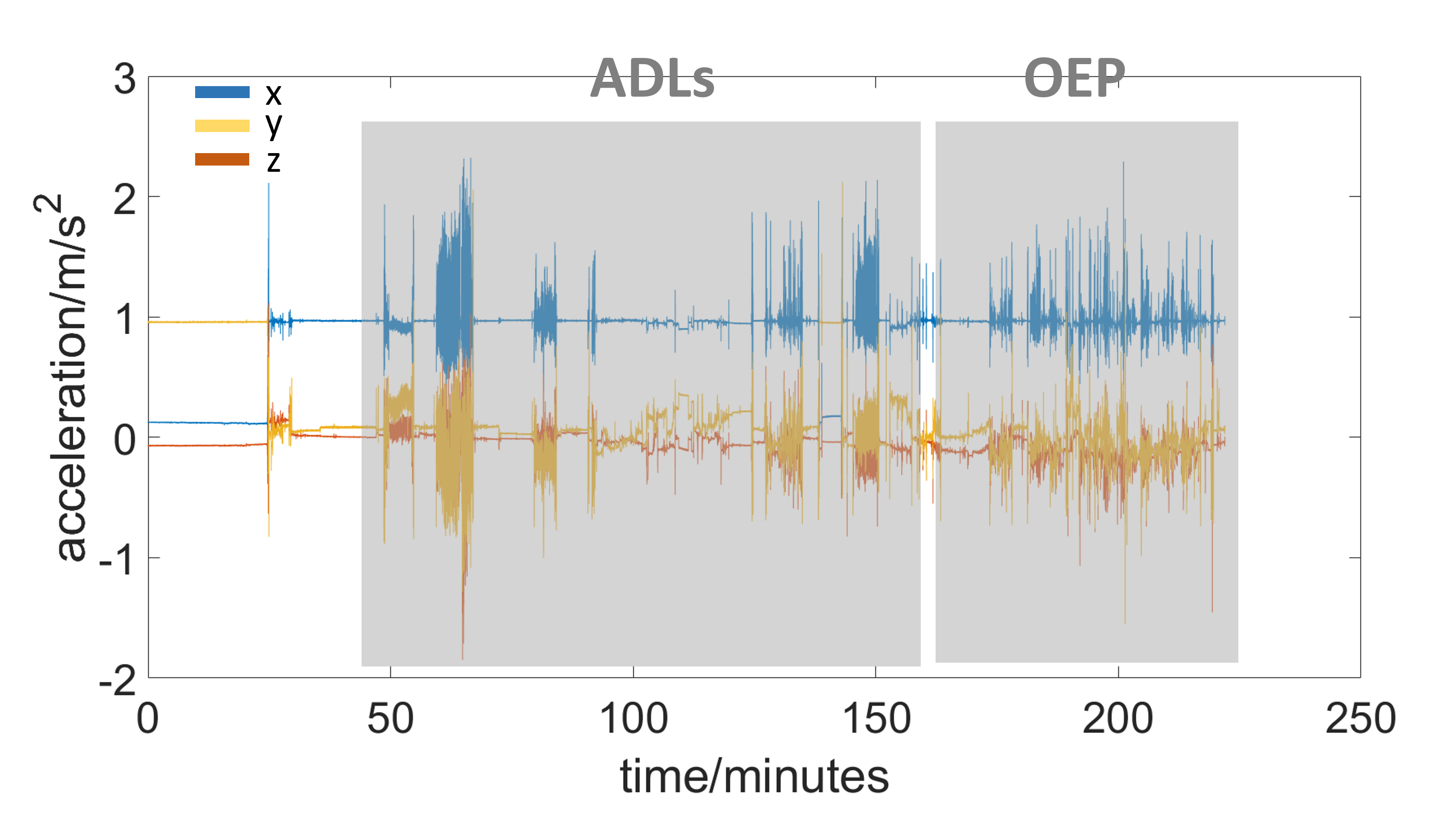}
\caption{An example of recorded acceleration from one subject}
\label{fig:signal}
\end{figure}

\begin{figure*}[!t]
\centering
\includegraphics[width=1.5\columnwidth]{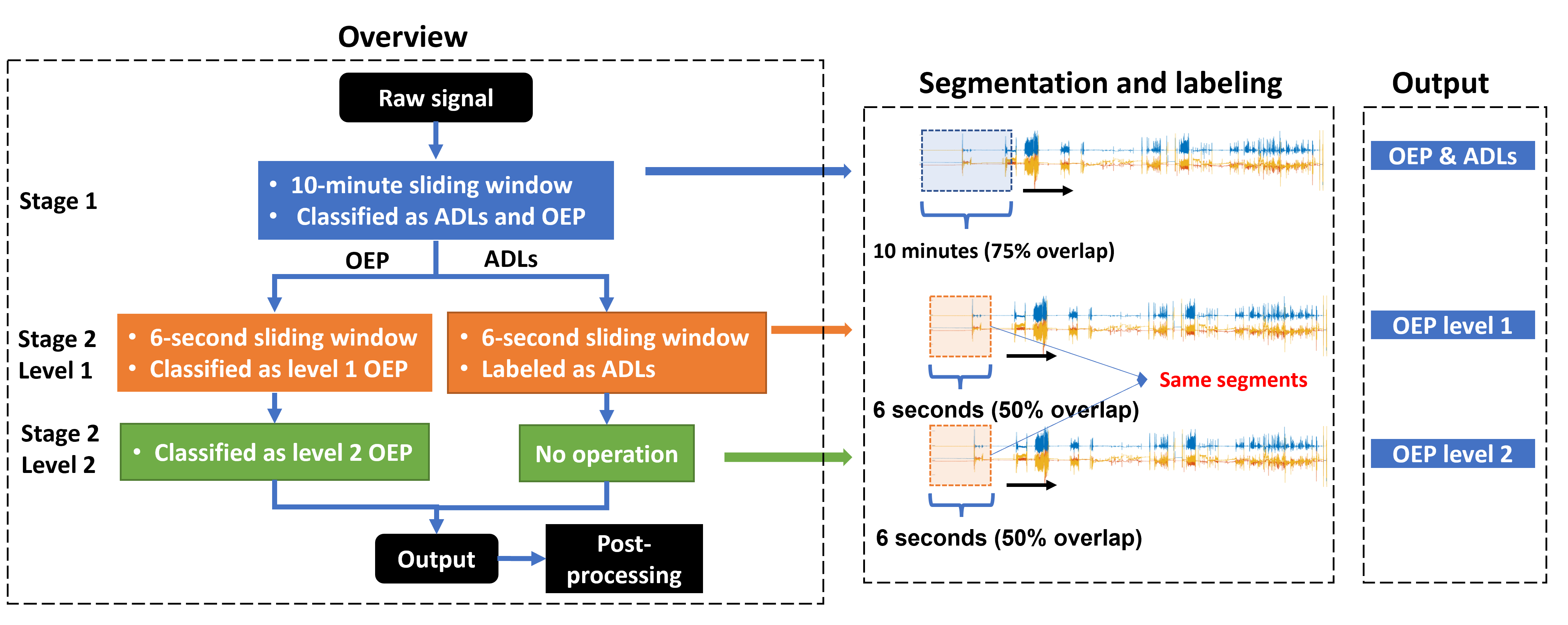}
\caption{An overview of the proposed system. NOTE: To better illustrate the process, the ratio of the sliding windows and signals in the figure does not correspond to the actual size. The post-processing stage is explained in the following sections.}
\label{overview}
\end{figure*}

An example of collected signals from a single subject is shown in Fig.~\ref{fig:signal}. The general overview of the two proposed experiments is shown in Fig.~\ref{overview}. Two stages of classification were proposed for different activities to be recognized. 

\subsubsection{Stage 1 (OEP vs. ADLs)}
Considering the characteristics of OEP and ADLs, a large sliding window size was applied to segment the signals. The optimal window size and overlap rate were identified by comparing performance across window sizes of 5, 10, and 15 minutes, and overlap rates of 25\%, 50\%, 75\%, and 80\%, using a CNN-BiLSTM model. The results are shown in section~\ref{subsec:resultsstage1} where a 10-minute window with 75\% overlap was selected. The segmented signals were then classified as OEP or ADLs (binary classification).

\subsubsection{Stage 2- level 1 (six classes of OEP classification)}
The signals were segmented with a smaller sliding window. The optimal window size and overlap rate were identified by comparing performance across window sizes of 2, 4, 6, and 8 seconds, and overlap rates of 25\%, 50\%, 75\%, and 80\%, using a Random Forest model. The results are shown in section~\ref{subsec:resultsstage2} where a 6-second window with 50\% overlap was selected. If the signals were classified as OEP in stage 1, they were further classified as the six classes of level 1 OEP as shown in Table~\ref{tab:OEPlabels}. Otherwise, they kept the ADLs labels as in stage 1.

\subsubsection{Stage 2- level 2 (15 classes of OEP classification)}
Following the results from level 1 classification, \textit{general walking} or \textit{general standing} were further classified by other models, whereas the other segments remained the same as level 1. There were consequently 15 classes in level 2 as shown in Table~\ref{tab:OEPlabels}. In level 2, the labels segmented in level 1 were further classified. Therefore, there were not any segmentation procedures.

\subsection{Pre-processing}
In this study, only the accelerometer ($a_x$, $a_y$, $a_z$) and gyroscope ($g_x$, $g_y$, $g_z$) from the IMU were used for classification, since they are more widely used than magnetometers for HAR in health care \cite{wang_survey_2019}. Also, adding a magnetometer did not improve the results by comparing classification performance. \par

For noise reduction, the raw signals were low-pass filtered by a 6-order Butterworth filter with a cut-off frequency of 10Hz. Then the signals were segmented using sliding windows as explained in~\ref{subsec:systemoverview} (10-minute window with 75\% overlap for stage 1 and 6-second window with 50\% overlap for stage 2).\par

An additional magnitude acceleration channel $a_M$ \cite{6597218, esfahani2017pams} was extracted according to the formula:
\begin{equation}
a_M=\sqrt{a_x^2+a_y^2+a_z^2},
\end{equation}

\subsection{Hand-crafted features and feature selection}
Time-domain and frequency-domain features were extracted from the seven channels (six original channels and one magnitude acceleration channel).  The types of hand-crafted features are shown in Table~\ref{tab:features}. In total, 15 time-domain features and nine frequency-domain features were extracted from each channel, all implemented using TSFEL package \cite{barandas2020tsfel}. Additionally, the subject information was included as meta-features, including the age, gender, weight, height, and health condition (sarcopenia, pre-sarcopenia, or no sarcopenia defined by EWGSOP1). In \cite{maekawa2011unsupervised}, it was found that the subject information had an added value for the recognition performance.

\begin{table}[!t]
\caption{Hand-crafted features extracted from the IMU channels}
\label{tab:features}
\centering
\scriptsize
\begin{tabular}{ll}
\hline
\textbf{Time-domain}                                              & \textbf{Frequency-domain}                                                    \\ \hline
Interquartile range                                               & FFT mean coefficient                                                         \\
Kurtosis                                                          & Fundamental frequency                                                        \\
Max                                                               & Human range energy (0.6-2.5Hz)*                                              \\
Mean                                                              & Max power spectrum                                                           \\
Median                                                            & Maximum frequency                                                            \\
Min                                                               & Median frequency                                                             \\
Root mean square                                                  & Spectral entropy                                                             \\
Skewness                                                          & Spectral kurtosis                                                            \\
Standard deviation                                                & Spectral skewness                                                            \\
Variance                                                          &                                                                              \\
Absolute energy                                                   &                                                                              \\
Autocorrelation                                                   &                                                                              \\
Centroid                                                          &                                                                              \\
Entropy                                                           &                                                                              \\
Zero crossing rate                                                &                                                                              \\ \hline
\multicolumn{2}{l}{\begin{tabular}[c]{@{}l@{}}*The frequency band was proposed as the dominant \\ frequency of human activity \cite{mannini_activity_2013}\end{tabular}}
\end{tabular}
\end{table}

Considering that some of the OEP exercises had to be performed with a certain order in the program (e.g. marching was always at the beginning of the OEP as a warm-up exercise), a feature named relative start time was extracted for each segment in stage 2:

\begin{equation}
f_{rst} = \frac{t_{sseg}-t_{sOEP}}{t_{eOEP}-t_{sOEP}},
\end{equation}

where $t_{sseg}$ denotes the start time of the segment (in stage 2), $t_{sOEP}$ denotes the predicted start time of the whole OEP, and $t_{eOEP}$ denotes the predicted end time of the whole OEP. From stage 1, $t_{sOEP}$ and $t_{eOEP}$ could be predicted. This feature indicates the approximate position of a certain exercise that occurred in the program.\par

In summary, 173 and 174 features were extracted for each segment in stage 1 and stage 2, respectively, since $f_{rst}$ was only extracted in stage 2.\par

Feature selection was performed by combining neighborhood component analysis and forward selection \cite{zhang2020bathroom}. After splitting training and validation sets, a feature set was selected for each training set, as the details explained in Section~\ref{datasplit}.

\subsection{Machine learning models}
\subsubsection{Classical machine learning}
Three classical models were applied: K-Nearest Neighbors (KNN), Support Vector Machines (SVM) with Radial Basis Function (RBF) kernel, and Random Forest (RF). Based on (standardized) hand-crafted features, these models were widely applied for HAR \cite{wang_survey_2019, nurwulan2020random}. They were applied in both stage 1 and stage 2. In stage 1, however, they were only applied as baseline models, since deep learning models were also applied.

\subsubsection{Deep learning}
Three deep learning models were applied in stage 1 classification: CNN, Transformer, and CNN-BiLSTM. They were not involved in stage 2 due to the small number of training examples. To reduce the computational cost, the input time series was down-sampled from 100Hz to 20Hz. Therefore, the input size to the models was 12000*6, where 12000 was the number of samples in 10 minutes and 6 was the number of original IMU channels.

\paragraph{CNN}
The CNN architecture was introduced in \cite{lee_human_2017}. It included three convolutional layers, each followed by a max-pooling layer to reduce the number of time samples and a dropout layer to avoid over-fitting. Then, the extracted features were flattened and classified by two fully-connected layers.

\paragraph{Transformer}
The Transformer architecture was proposed by \cite{dirgova2022wearable}. After normalization and position embedding, three encoder blocks were applied. Each encoder block consisted of a multi-head attention layer, a fully-connected layer, and dropout layers. Then, the classification results were generated by another fully-connected layer.

\begin{figure}[!t]
\centering
\includegraphics[width=0.6\columnwidth]{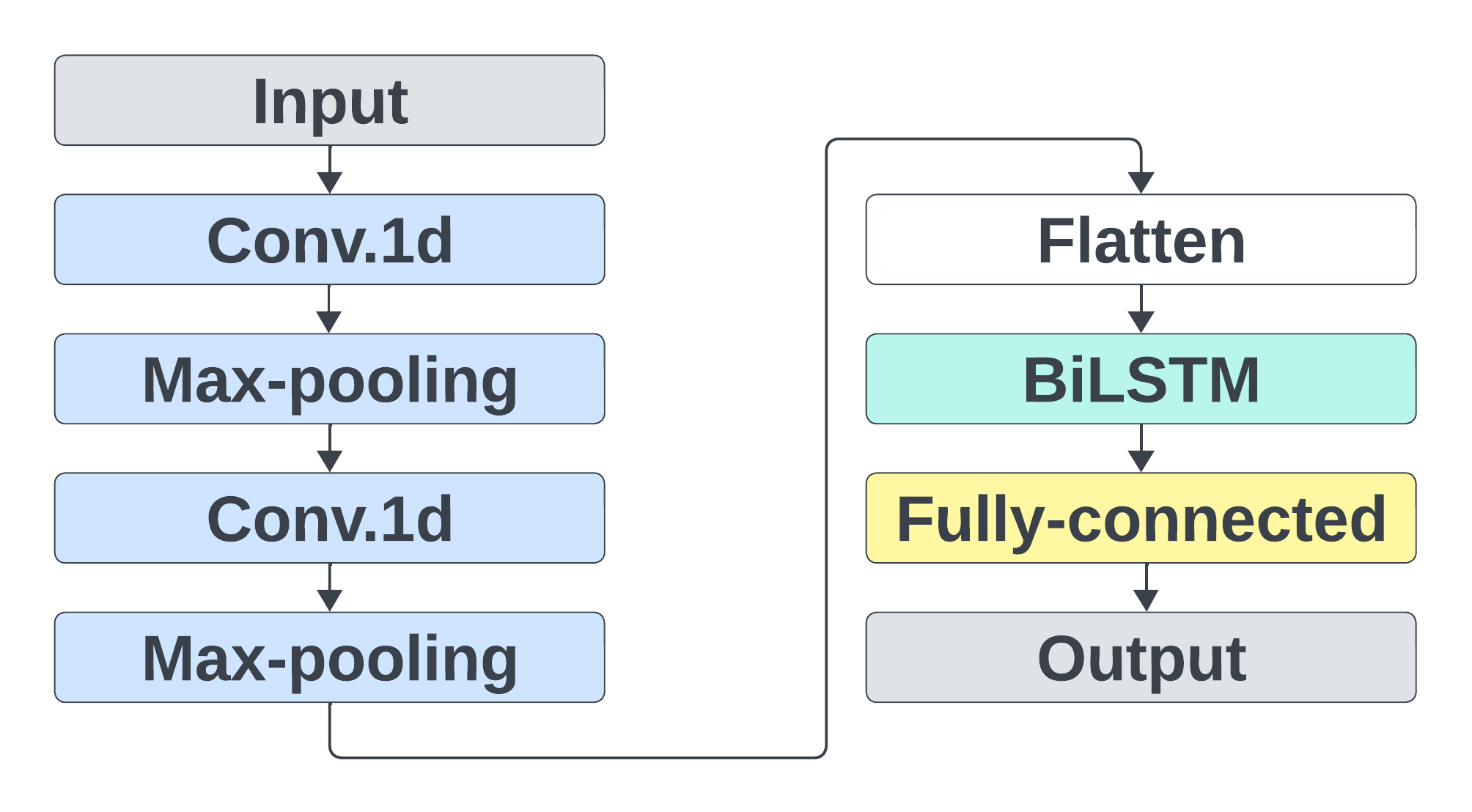}
\caption{The architectures of the CNN-BiLSTM model. Each convolutional layer and LSTM layer was followed by a dropout layer, which is not shown in the figure. The hyperparameters of were tuned according to Table~\ref{tab:hyper}.}
\label{fig:DL}
\end{figure}

\paragraph{CNN-BiLSTM}
The CNN-BiLSTM model applied two convolutional layers for local features learning and a following BiLSTM layer for temporal features learning, as shown in Fig.~\ref{fig:DL}.

The deep learning models were built on tensorflow 2.8 and were trained on a NVIDIA P100 SXM2 GPU. A batch size of 64 was applied for training. Adam algorithm was used to optimize the cross-entropy loss function. To reduce the influence of the imbalanced dataset, all models applied the "class weight" option to weight the loss function according to the number of samples for each class. \par

\begin{figure}[!t]
\centering
\includegraphics[width=0.8\columnwidth]{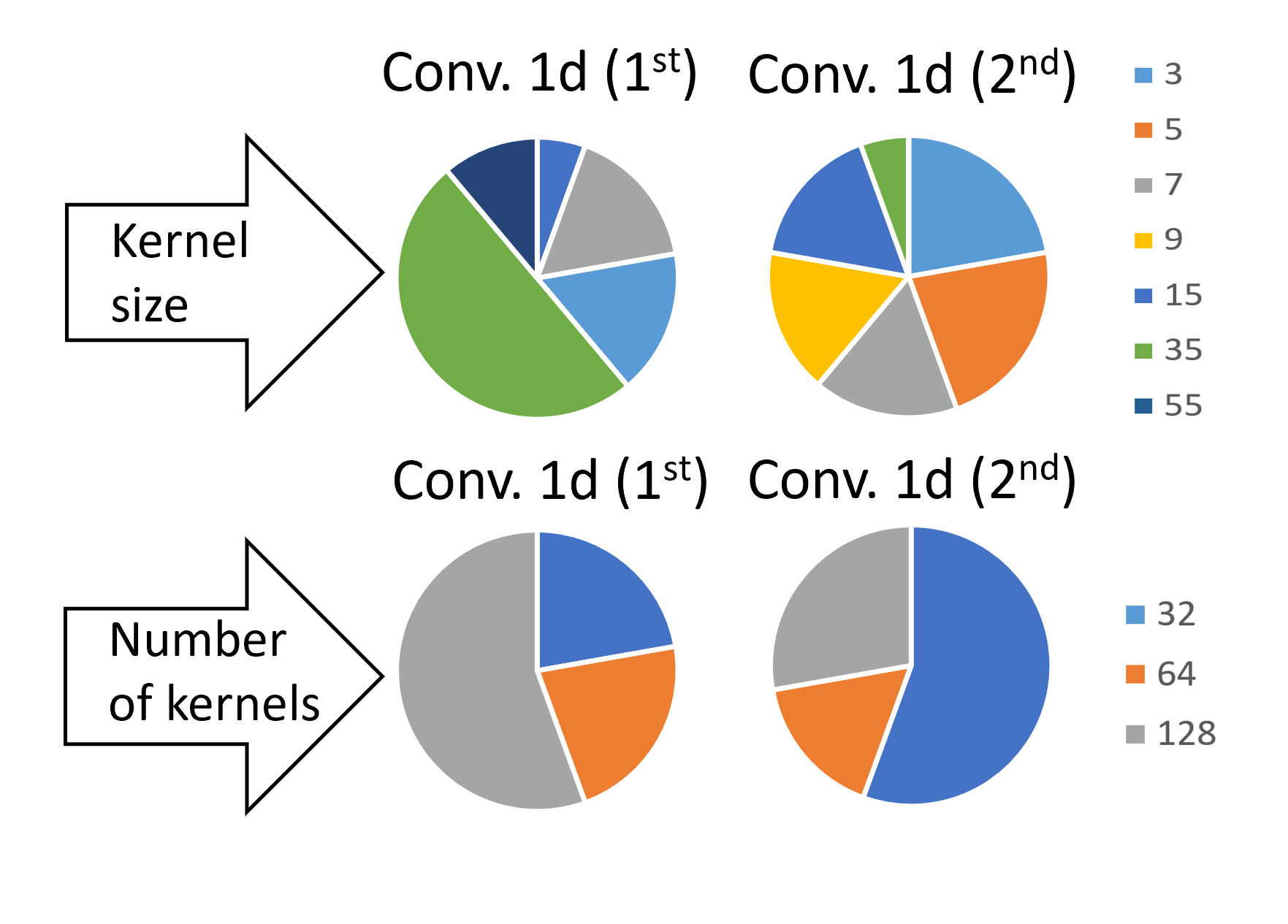}
\caption{Some selected hyperparameters in each outer loop from CNN-BiLSTM in stage 1 (dataset 1+ dataset 2).}
\label{fig:hyper}
\end{figure}

The hyperparameters were tuned using Hyperband Search \cite{li2017hyperband}. The search spaces of the hyperparameters for each model are shown in Table~\ref{tab:hyper}. Since the input time series was large in stage 1 (input length= 12000 samples), the kernel size in the convolutional layers was tuned in a large search space. Fig.~\ref{fig:hyper} shows some selected hyperparameters in each outer loop from CNN-BiLSTM in stage 1. The test and validation of the models are described in Section~\ref{datasplit}. 

\begin{table}[!t]
\centering
\caption{The search space of each tuned hyperparameter of the models}
\label{tab:hyper}
\begin{tabular}{|l|l|l|}
\hline
\textbf{Model}                                                                   & \textbf{Hyper-parameter}                                                     & \textbf{Search space}        \\ \hline
\multirow{2}{*}{SVM}                                                             & \begin{tabular}[c]{@{}l@{}}Regularization \\ parameter\end{tabular}          & [10, 1, 0.1, 0.01, 0.001]    \\ \cline{2-3} 
                                                                                 & \begin{tabular}[c]{@{}l@{}}Kernel value\\ (gamma)\end{tabular}               & [10, 1, 0.1, 0.01, 0.001]    \\ \hline
KNN                                                                              & \begin{tabular}[c]{@{}l@{}}Number of \\ neighbors\end{tabular}               & [3, 5, 7, 9]                 \\ \hline
\multirow{2}{*}{RF}                                                              & Number of trees                                                              & [50, 100, 200]               \\ \cline{2-3} 
                                                                                 & \begin{tabular}[c]{@{}l@{}}Maximum depth \\ of the tree\end{tabular}         & [1:1:35]                     \\ \hline
\multirow{8}{*}{\begin{tabular}[c]{@{}l@{}}Deep learning \\ models\end{tabular}} & \begin{tabular}[c]{@{}l@{}}Conv. 1d:\\ Kernel size\end{tabular}              & [3, 5, 7, 9, 11, 15, 35, 55] \\ \cline{2-3} 
                                                                                 & \begin{tabular}[c]{@{}l@{}}Conv. 1d:\\ Number of filters\end{tabular}        & [32, 64, 128]                \\ \cline{2-3} 
                                                                                 & \begin{tabular}[c]{@{}l@{}}BiLSTM:\\ Number of units\end{tabular}            & [32, 64, 128]                \\ \cline{2-3} 
                                                                                 & \begin{tabular}[c]{@{}l@{}}Attention:\\ Number of heads\end{tabular}         & [3, 4, 5, 6]                 \\ \cline{2-3} 
                                                                                 & \begin{tabular}[c]{@{}l@{}}Fully-connected:\\ Number of neurons\end{tabular} & [32, 64, 128]                \\ \cline{2-3} 
                                                                                 & Droupout rate                                                                & [0.1, 0.2, 0.5]              \\ \cline{2-3} 
                                                                                 & Learning rate                                                                & [0.1, 0.01, 0.001, 0.0001]   \\ \cline{2-3} 
                                                                                 & Pooling size                                                                 & [2, 3, 5, 10]                \\ \hline
\end{tabular}
\end{table}

\subsection{Post-processing}
The transition activities and ADLs could be similar to the OEP sub-classes. For example, \textit{sit-to-stand} also happened in transition activities and ADLs. However, only consecutive \textit{sit-to-stand} labels should be classified as OEP sub-classes. To reduce the negative influence of the ADLs and transition activities in stage 1 and stage 2, post-processing was applied to improve recognition performance. \par

To smooth the predicted labels O = $\{o_{1}, o_{2},… , o_{n}\}$, as a time series, the post-processing algorithm was applied, as described in Algorithm~\ref{alg:ALG1}. It took the predicted segments as input and returned smoothed segments P = $\{p_{1}, p_{2},… , p_{n}\}$. A smoothing window was moving along the time series. The values of post-processing window lengths were selected from a range of values (from 10 to 20 minutes for stage 1, and from 21 to 39 seconds for stage 2) by comparing classification performance. Finally, the window length was seven segments (i.e. 17.5 minutes) for stage 1, and 11 segments (i.e. 33 seconds) for stage 2. The label was classified as transition activities if the most adjacent labels were from different classes. The algorithm could correct some misclassified segments, which is shown in section~\ref{subsec:resultsstage1}.

\begin{algorithm}[!t]
\caption{Post-processing}
  \label{alg:ALG1}
\begin{algorithmic}
\STATE \textbf{INPUT:} a time series of the original predicted labels O = $\{o_{1}, o_{2},… , o_{n}\}$; k=3 for stage 1 and k=5 for stage 2
\STATE \textbf{OUTPUT:} a time series of the post-processed predicted labels P = $\{p_{1}, p_{2},… , p_{n}\}$
\STATE
\STATE  Initialize P with the same length as O
\FOR{i=k+1, k+2, k+3, ..., n-k-2, n-k-1, n-k}
       \STATE $p_{i}$ = the majority class of the sequence $\{o_{t}\in O | (i-k)\leq t\leq (i+k)\}$
\ENDFOR      

\end{algorithmic}
\end{algorithm}

\subsection{Dataset split}
\label{datasplit}

The validation for dataset 1 applied nested Leave-One-Subject-Out Cross-Validation (LOSOCV). For each outer loop, the data from one subject was in the test set whereas the rest was in the training set. For hyperparameters tuning, in each inner loop, a single subject was excluded from the validation set. In other words, the best average weighted f1-scores of all validation sets in the inner loops were obtained. The best model hyperparameters and the best feature set were then applied to the test set in the outer loop. The final evaluation performance was calculated from all test sets.\par

The validation for dataset 2 was similar to dataset 1. The only difference is that dataset 1 was also included in the training set, to further improve the classification performance.

\subsection{Evaluation}
\subsubsection{window-wise evaluation}

For each class c, the f1-score was calculated as:
\begin{equation}
f1_{c}=2*\frac{precision_{c}*recall_{c}}{precision_{c}+recall_{c}},
\label{f:f1}
\end{equation}

where
\begin{equation}
precision_{c}=\frac{TP_{c}}{TP_{c}+FP_{c}}, and
\label{f:precision}
\end{equation}
\begin{equation}
recall_{c}=\frac{TP_{c}}{TP_{c}+FN_{c}}.
\label{f:recell}
\end{equation}

where $TP_{c}$, $FP_{c}$, and $FN_{c}$ denote the numbers of true positive, false positive, and false negative labels.\par

For overall evaluation in stage 2, weighted f1-score was calculated from the f1-score and the number of labels of class c:
\begin{equation}
weighted\:f1=\sum_{c}f1_{c}*N_{c},
\end{equation}

where N$_{c}$ denotes the number of true labels belonging to class c. 

\subsubsection{segment-wise evaluation}

For this evaluation method, the predicted labels based on the sliding windows were reconstructed as a time series. A segment was defined as an aggregation of consecutive labels that belonged to the same class \cite{farha_ms-tcn_2019}. Then segmental f1-scores of Intersection over Union (IoU) were calculated for each class. The IoU matrix was defined as the ratio of intersection and the union of the predicted and true segments. After the true and predicted labels were reconstructed as a time series, a threshold of IoU values was set. Then, segment-wise true positive ($TPseg$), false positive ($FPseg$), and false negative ($FNseg$) were defined:

\begin{itemize}
	\item $TPseg$: IoU$\geq$threshold
	\item $FPseg$: IoU$<$threshold, true segments shorter than predicted segments
	\item $FNseg$: IoU$<$threshold, true segments longer than predicted segments
\end{itemize}
	
\begin{figure}[!t]
\centering
\includegraphics[width=\columnwidth]{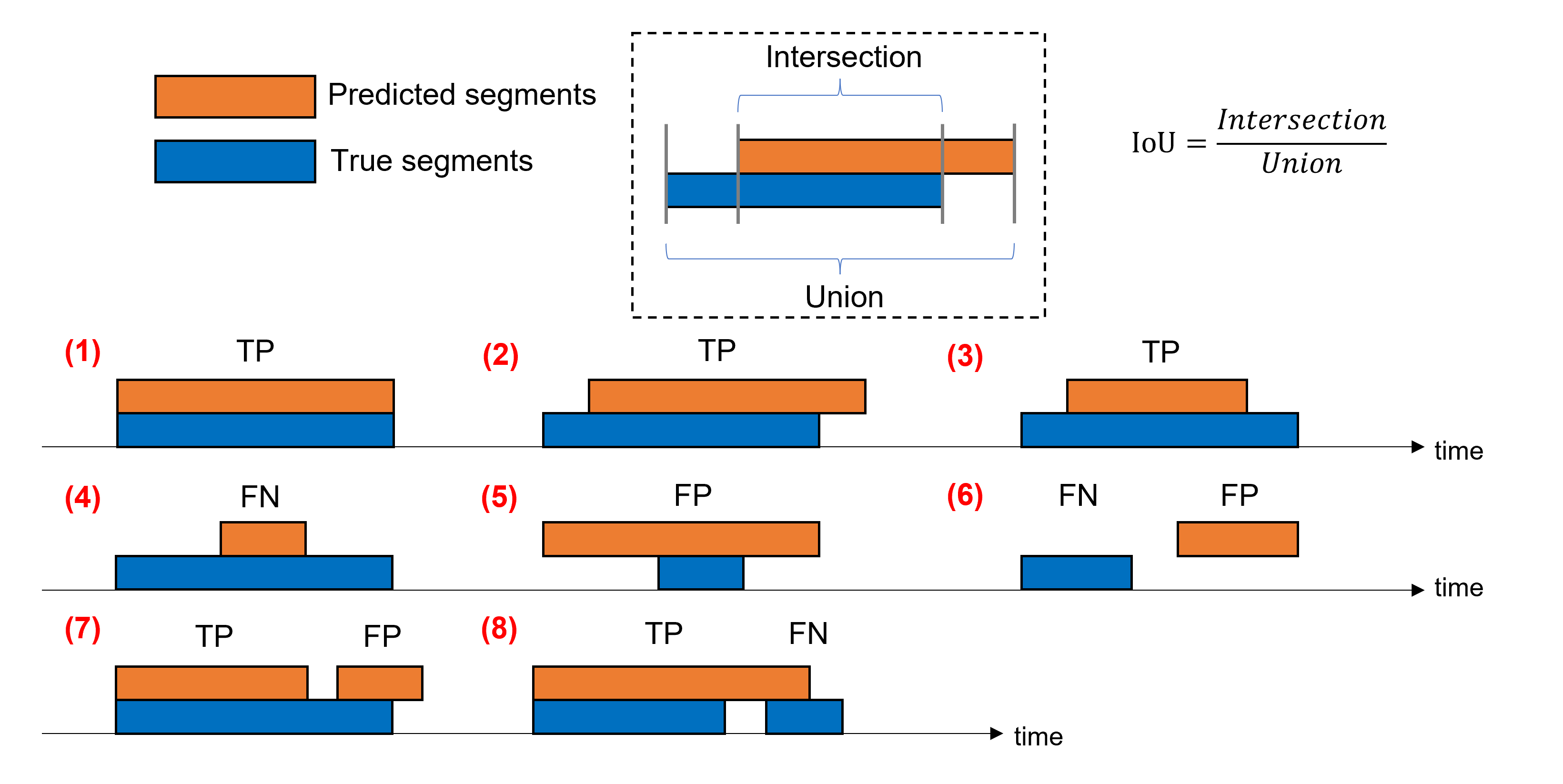}
\caption{The definition of IoU, TP, FP, and FN. There are eight cases shown in the figure. In case 4 and case 5, FN or FP depends on the length of the true and predicted segment. In case 7, if a true segment is predicted as some smaller segments, FP numbers increase. In case 8, if some separate true segments are predicted as one segment, FN numbers increase.}
\label{fig:IoU}
\end{figure}

Fig.~\ref{fig:IoU} illustrates the definition of IoU, TP, FP, and FN. From these values, segment-wise precision, recall, and f1-scores could be calculated by formula~\ref{f:f1},~\ref{f:precision},~\ref{f:recell}. Compared with the window-wise evaluation, this method could also evaluate the over-segmentation error. In this study, the overlaps of 0.5 and 0.75 were applied for stage 1 evaluation.

\section{Results}
\label{sec:results}

\subsection{Stage 1 classification}
\label{subsec:resultsstage1}

For both datasets, a 10-minute sliding window was selected from 5, 10, and 15 minutes, and an overlap rate of 75\% was selected from 25\%, 50\%, 75\%, and 80\%, using a CNN-BiLSTM model. They were hence applied for stage 1 classification. Table~\ref{tab:comparef1stage1} shows the impacts of some values of window size and overlap rate. 

\begin{table}[!t]
\centering
\caption{OEP f1-scores using different sliding window sizes (overlaps) in stage 1 using CNN-BiLSTM}
\label{tab:comparef1stage1}
\begin{tabular}{llll}
\hline
          & 5-min (50\%)    & 10-min (75\%)   & 15-min (85\%)          \\ \hline
dataset 1 & 0.924 & \textbf{0.983} & 0.960  \\ \hline
dataset 2 & 0.916 & \textbf{0.950} & 0.943 \\ \hline
\end{tabular}
\end{table}

The window-wise and segment-wise f1-scores of OEP classified by different machine learning models are shown in Table.~\ref{tab:stage1f1}. The window-wise f1-scores of the CNN-BiLSTM architecture were higher than the other models in both datasets (0.983 and 0.950, respectively). Although the f1-scores dropped from dataset 1 to dataset 2, such a drop is smaller using the CNN-BiLSTM models. For dataset 2, the IoU f1-scores of classical machine learning models significantly decreased. The IoU f1-scores (75\%) of CNN-BiLSTM were found to be 0.867 and 1.000, respectively, which outperformed the other models.\par

\begin{table}[]
\centering
\footnotesize
\caption{The window-wise (f1, precision, recall) and segment-wise (IoU f1) evaluation in stage 1 by different models after post-processing}
\label{tab:stage1f1}
\begin{tabular}{lllllll}
\hline
                                 & KNN   & RF    & SVM   & CNN   & \begin{tabular}[c]{@{}l@{}}Trans-\\ former\end{tabular} & \begin{tabular}[c]{@{}l@{}}CNN-\\ BiLSTM\end{tabular} \\ \hline
\multicolumn{7}{c}{dataset 1}                                                                                                                                                      \\ \hline
\multicolumn{1}{l|}{f1}          & 0.869 & 0.949 & 0.847 & 0.874 & 0.949                                                   & \textbf{0.983}                                        \\ \cline{2-7} 
\multicolumn{1}{l|}{precision}   & 0.952 & 0.984 & 0.959 & 0.862 & 0.984                                                   & \textbf{1.000}                                        \\ \cline{2-7} 
\multicolumn{1}{l|}{recall}      & 0.799 & 0.916 & 0.759 & 0.887 & 0.916                                                   & \textbf{0.967}                                        \\ \hline
\multicolumn{1}{l|}{IoU f1@0.5}  & 0.938 & 1.000 & 1.000 & 0.714 & 0.762                                                   & \textbf{1.000}                                        \\ \cline{2-7} 
\multicolumn{1}{l|}{IoU f1@0.75} & 0.750 & 0.133 & 0.133 & 0.619 & 0.654                                                   & \textbf{0.867}                                        \\ \hline
\multicolumn{7}{c}{dataset 2}                                                                                                                                                      \\ \hline
\multicolumn{1}{l|}{f1}          & 0.616 & 0.692 & 0.627 & 0.849 & 0.909                                                   & \textbf{0.950}                                        \\ \cline{2-7} 
\multicolumn{1}{l|}{precision}   & 0.507 & 0.585 & 0.571 & 0.789 & 0.927                                                   & \textbf{0.931}                                        \\ \cline{2-7} 
\multicolumn{1}{l|}{recall}      & 0.786 & 0.847 & 0.694 & 0.918 & 0.891                                                   & \textbf{0.969}                                        \\ \hline
\multicolumn{1}{l|}{IoU f1@0.5}  & 0.000 & 0.087 & 0.000 & 0.857 & 0.821                                                   & \textbf{1.000}                                        \\ \cline{2-7} 
\multicolumn{1}{l|}{IoU f1@0.75} & 0.000 & 0.087 & 0.000 & 0.857 & 0.125                                                   & \textbf{1}                                            \\ \hline
\end{tabular}
\end{table}

\begin{figure}[!t]
\centering
\includegraphics[width=\columnwidth]{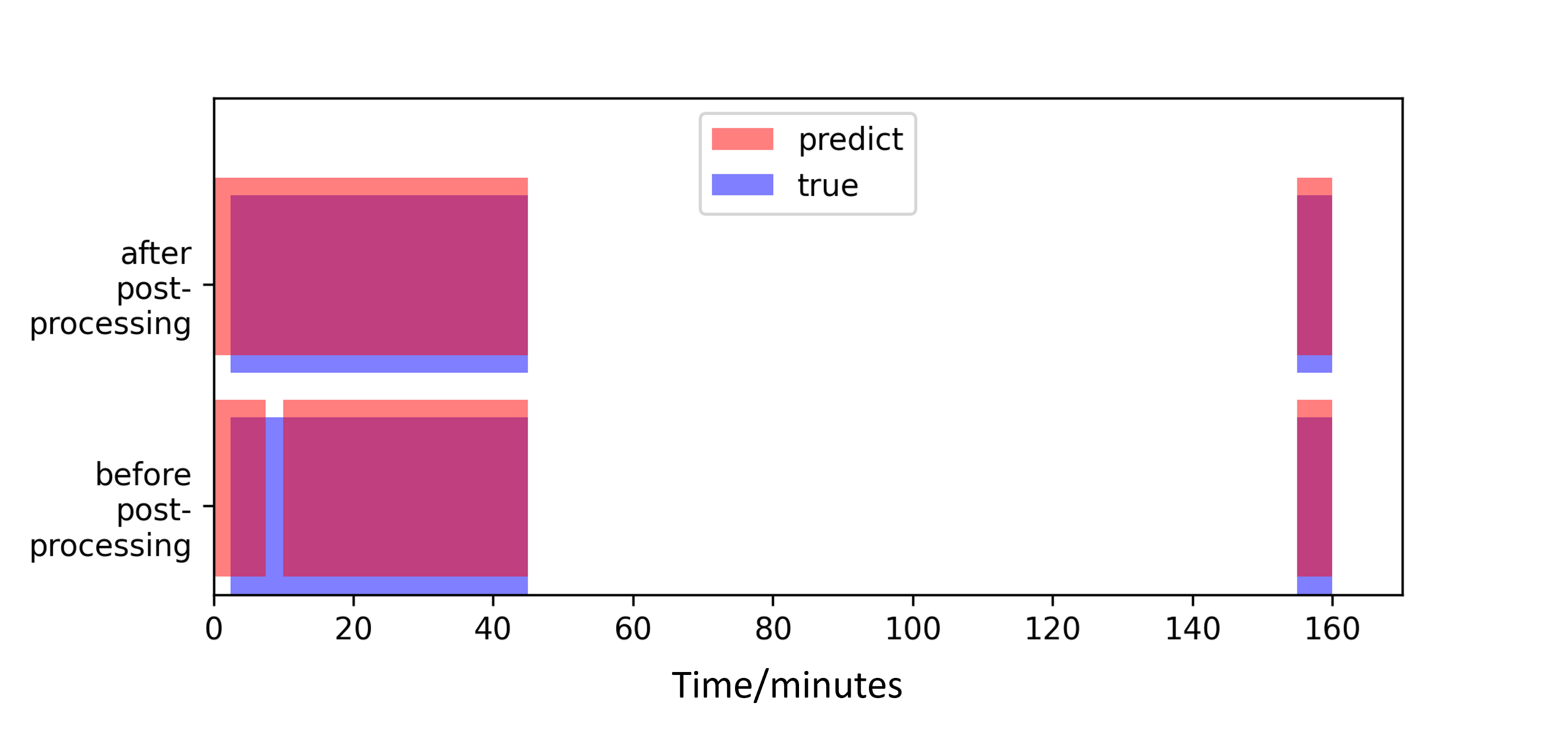}
\caption{An example of reconstructed time-series of labels}
\label{fig:segment-wise1}
\end{figure}

An example of the predicted time series of stage 1 is shown in Fig~\ref{fig:segment-wise1}, which shows the impacts of post-processing. \par

\begin{figure}[!t]
\centering
\includegraphics[width=0.8\columnwidth]{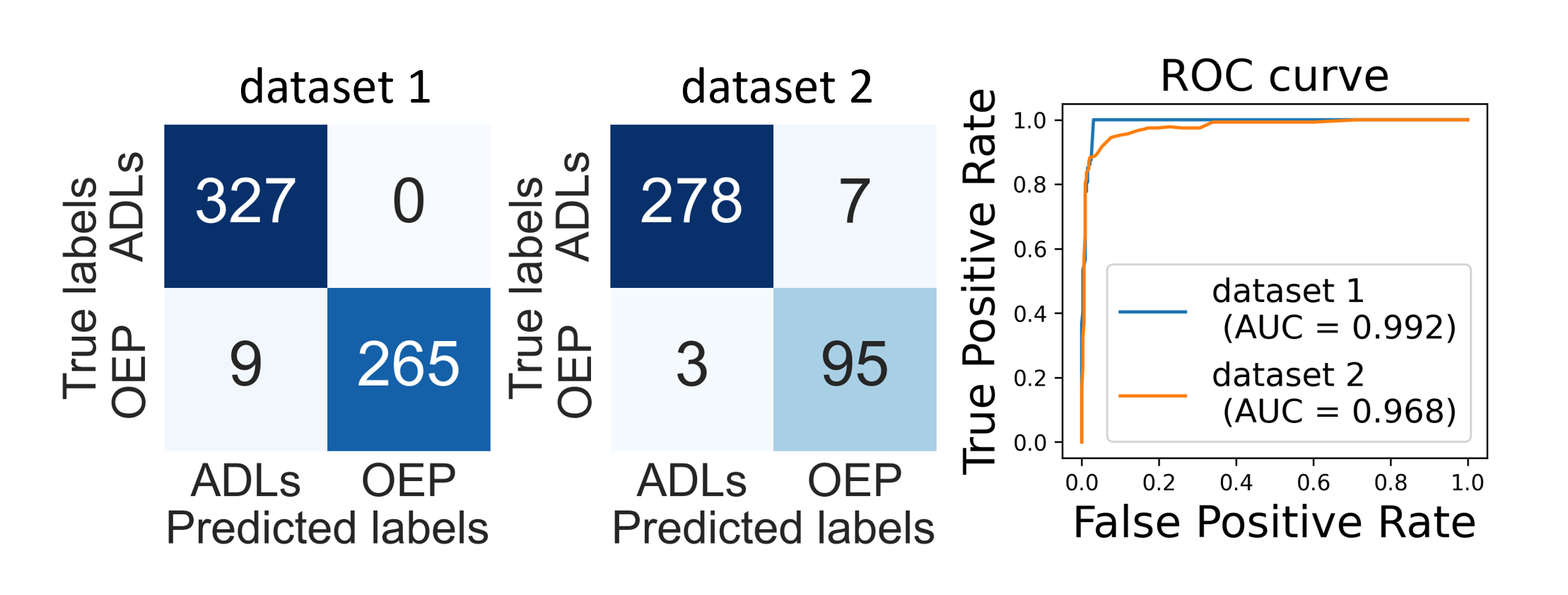}
\caption{Confusion matrices and ROC curves for stage 1 classification by CNN-BiLSTM after post-processing}
\label{fig:level0cm}
\end{figure}

\begin{figure}[!t]
\centering
\includegraphics[width=0.7\columnwidth]{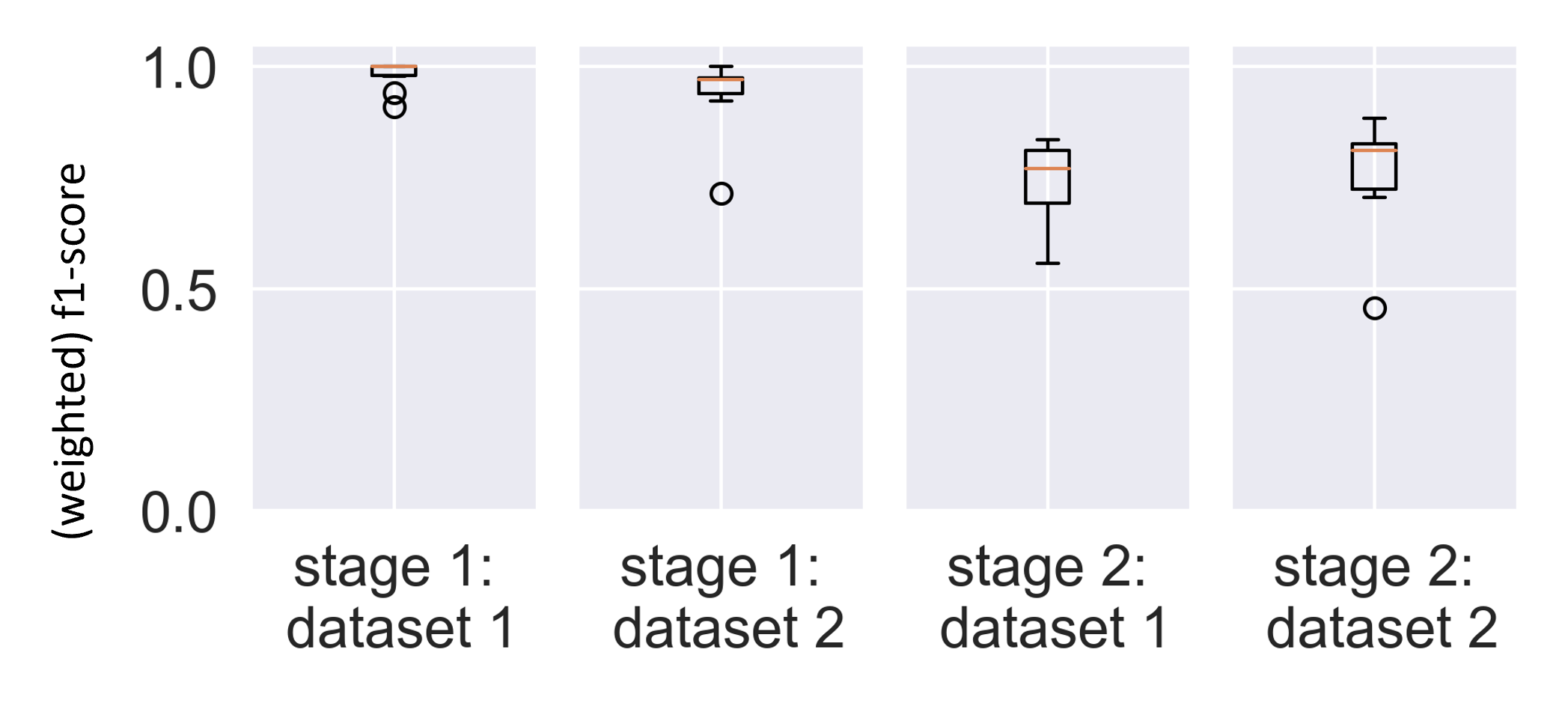}
\caption{Boxplots of the f1-scores in stage 1 and weighted f1-scores in stage 2 for each subject}
\label{fig:box}
\end{figure}

The confusion matrices and receiver operating characteristic (ROC) curves by the CNN-BiLSTM models are shown in Fig.~\ref{fig:level0cm}. For dataset 1, all the ADLs segments were classified correctly. On the other hand, nine OEP segments were classified as ADLs. For dataset 2, seven ADLs segments were classified as OEP while three OEP segments were classified as ADLs segments. Fig.~\ref{fig:box} shows the boxplot illustrating the f1-scores for each subject in stage 1. According to the figure, there were three outliers in total. Apart from this, the f1-scores among subjects showed low variance.

\subsection{Stage 2 classification}
\label{subsec:resultsstage2}

For both datasets, a 6-second sliding window was selected from 2, 4, 6, and 8 seconds, and an overlap rate of 50\% was selected from 25\%, 50\%, 75\%, and 80\%, using a RF model. They were hence applied for stage 2 classification. Table~\ref{tab:comparef1stage1} shows the impacts of some values of window size and overlap rate.

\begin{table}[!t]
\centering
\caption{Weighted f1-scores using different sliding window sizes in stage 2 (with 50\% overlap) using RF}
\label{tab:comparef1stage2}
\begin{tabular}{lllll}
\hline
          & 2s    & 4s    & 6s             & 8s    \\ \hline
dataset 1 & 0.733 & 0.741 & \textbf{0.743} & 0.739 \\ \hline
dataset 2 & 0.722 & 0.775 & \textbf{0.798} & 0.766 \\ \hline
\end{tabular}
\end{table}

As feature selection was performed on each iteration in LOSOCV, each training set resulted in a different feature set. The ten most frequently selected features were: Relative start time, Entropy ($a_y$), Entropy ($a_M$), Entropy ($a_x$), Mean ($g_z$), Centroid ($a_M$), Median ($g_x$), Centroid ($g_x$), Kurtosis ($a_y$), Centroid ($a_x$).\par

\begin{figure}[!t]
\centering
\subfloat[Dataset 1]{\includegraphics[width=\columnwidth]{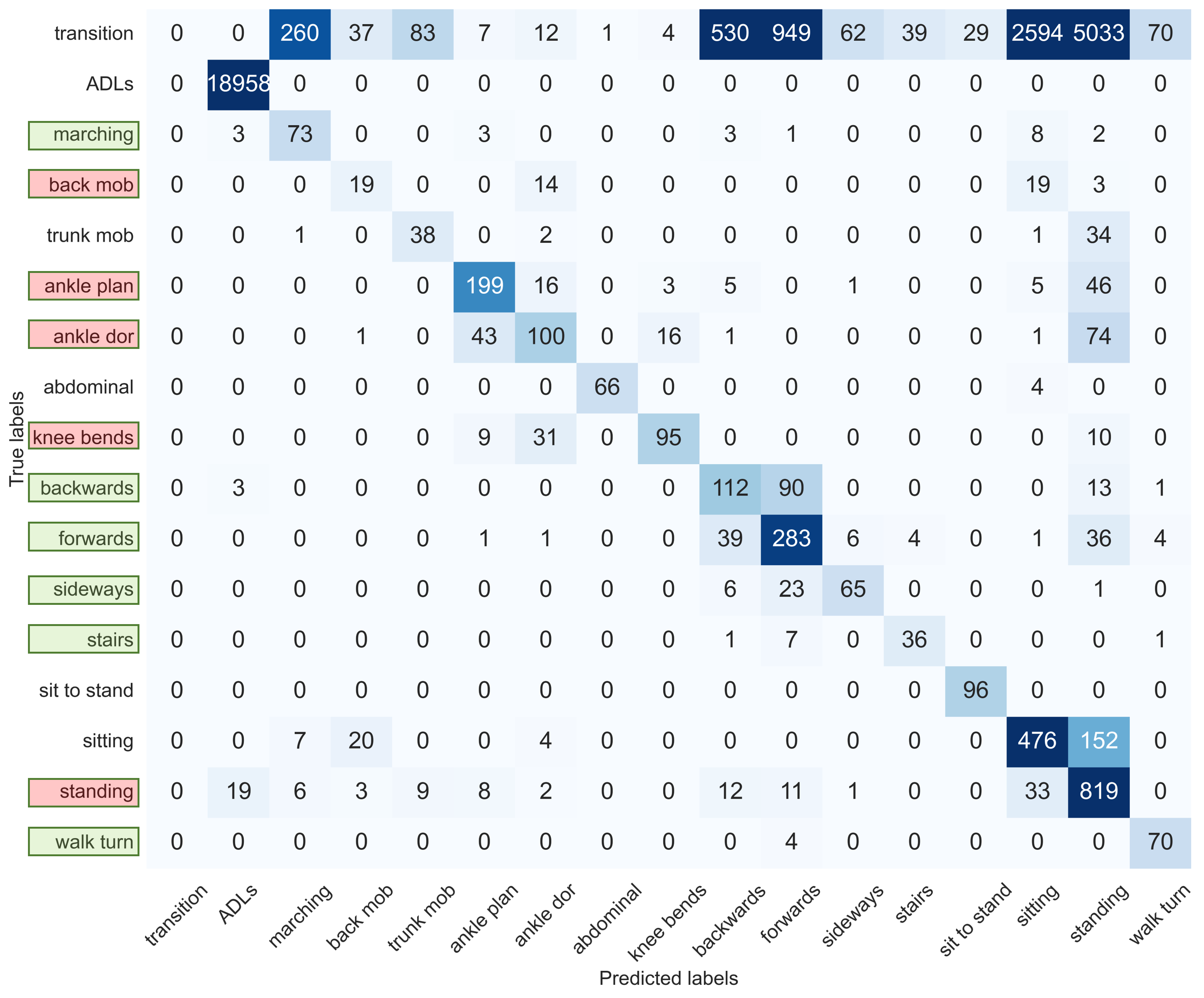}%
\label{fig:stage2cm1}}
\hfill
\subfloat[Dataset 2]{\includegraphics[width=\columnwidth]{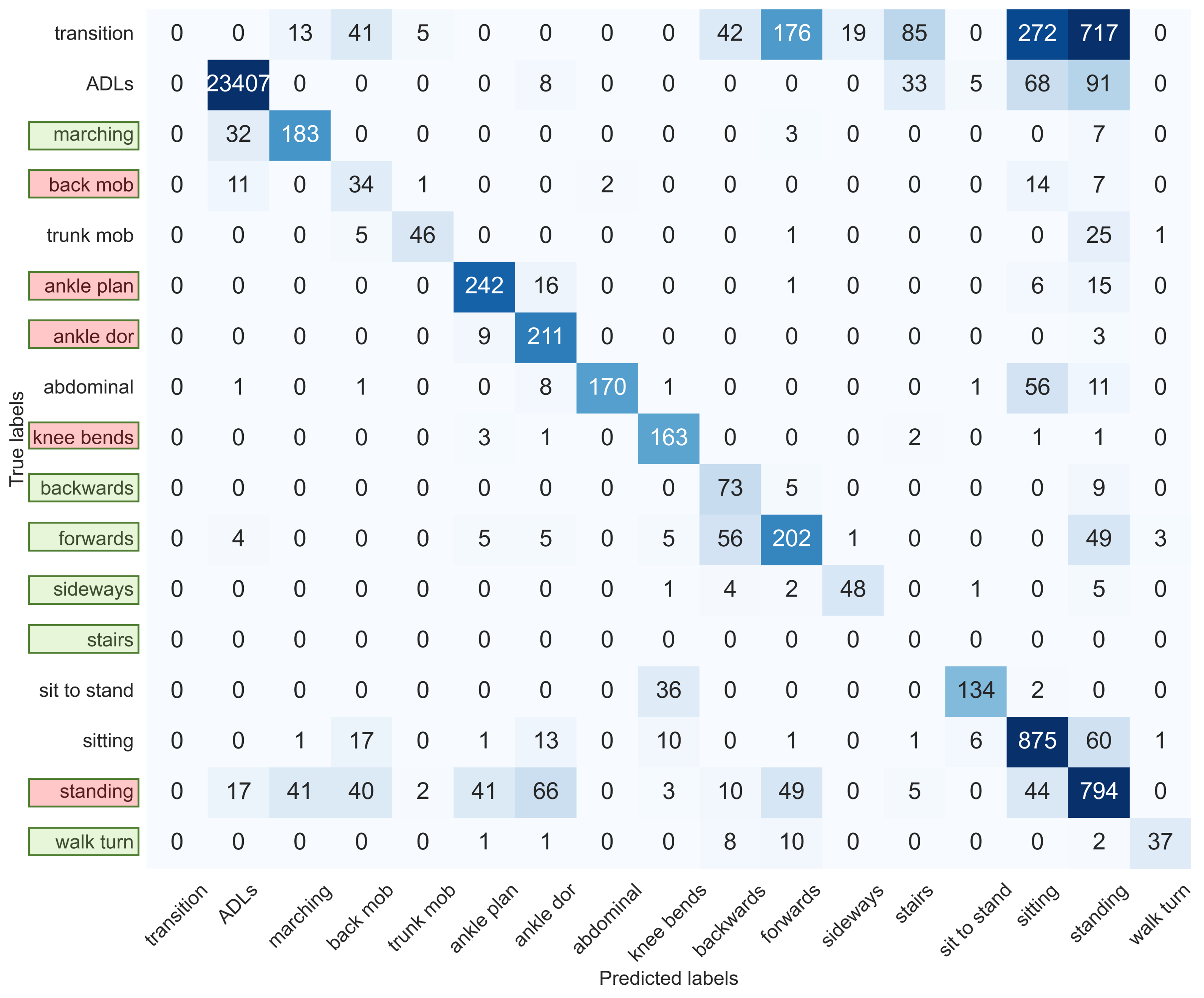}%
\label{fig:stage2cm2}}
\caption{Confusion matrices for stage 2 classification by RF after post-processing. The green labels denote \textit{general walking} and the red labels denote \textit{general standing} in level 1. (abbreviations: plan= plantarflexors, dor= dorsiflexors, mob= mobilizer). There are no FP labels for transition activities class, because the labels were not in the training set.}
\label{fig:stage2cm}
\end{figure}

The weighted f1-scores in stage 2 by different models are shown in Table~\ref{tab:level2weighted}. For both datasets, RF models obtained the best weighted f1-scores of 0.743 and 0.798, respectively. The confusion matrices for both datasets by RF are shown in Fig.~\ref{fig:stage2cm}. Compared with stage 1, ADLs in stage 2 showed less confusion because of the further post-processing in stage 2. On the other hand, the transition activities led to false positive predictions for most of the classes.

\begin{table}[!t]
\centering
\caption{Weighted f1-scores in stage 2 by different models}
\label{tab:level2weighted}
\begin{tabular}{p{1.5cm}p{1cm}p{1cm}p{1cm}}
\hline
          & KNN   & RF             & SVM   \\ \hline
dataset 1 & 0.653 & \textbf{0.743} & 0.689 \\
dataset 2 & 0.699 & \textbf{0.798} & 0.724 \\ \hline
\end{tabular}
\end{table}

\begin{figure}[!t]
\centering
\includegraphics[width=0.9\columnwidth]{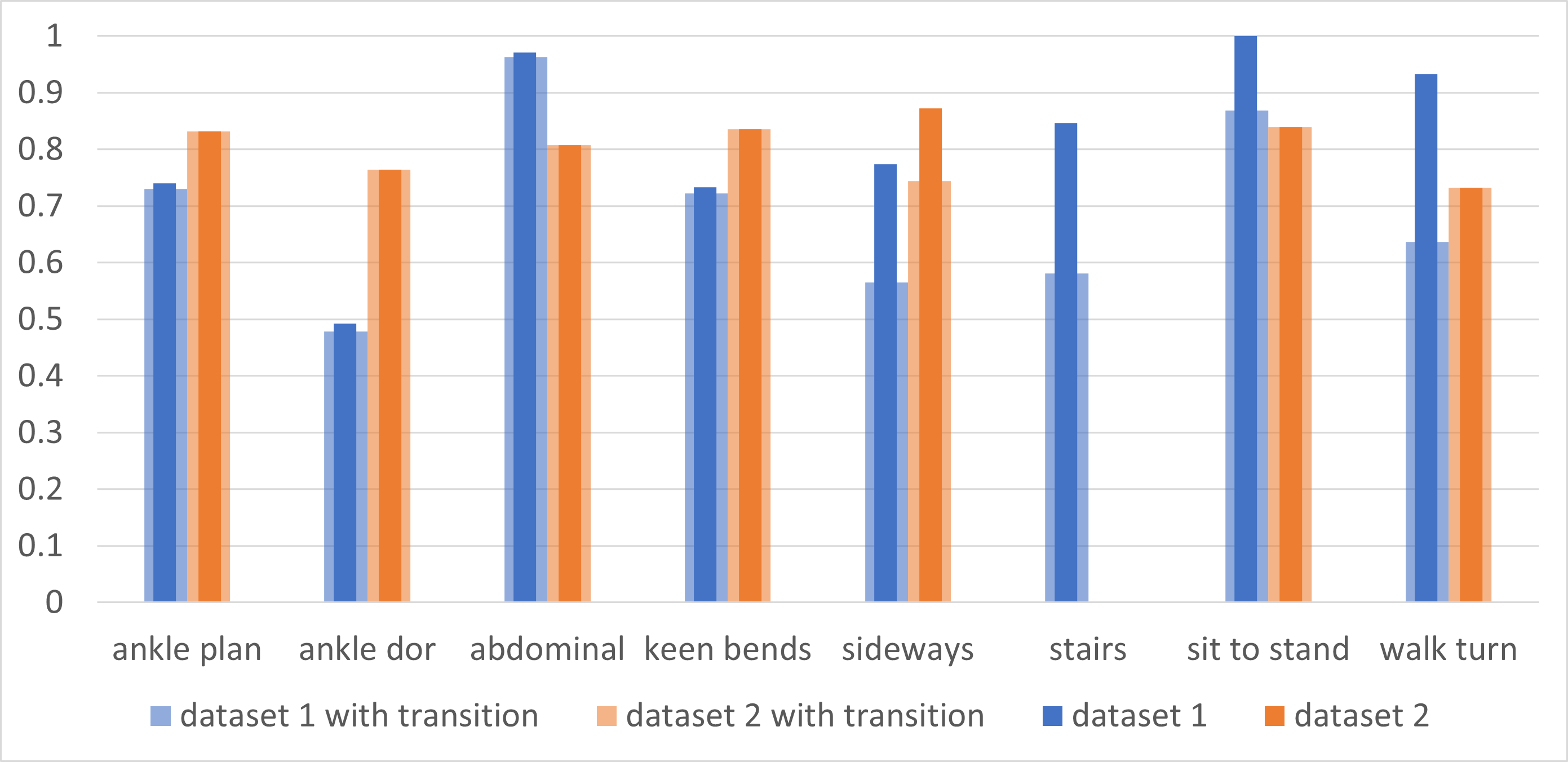}
\caption{F1-scores of some classes in stage 2 by RF after post-processing. The light colors illustrates the results with counting the transition activities.}
\label{fig:f1stage2}
\end{figure}

Fig.~\ref{fig:f1stage2} shows the f1-scores for some activities. As illustrated by the light colors, in dataset 2 at home, there were four activities beyond the threshold of 0.8: \textit{ankle plantarflexors}, \textit{abdominal muscles}, \textit{knee bends}, and \textit{sit-to-stands}. With transition classified as FP, there were three activities with f1-scores decreased by more than 0.1: \textit{sideways walking}, \textit{stairs walking}, and \textit{walking and turn}. Fig.~\ref{fig:box} shows the boxplot illustrating the weighted f1-scores for each subject in stage 2, showing larger variance than stage 1.

\section{DISCUSSION}
\label{sec:discussion}

\subsection{Stage 1}
This study reported high f1-scores for OEP recognition in both lab and home environments. In stage 1, a large sliding window size (10 minutes) was applied to segment the signals. Compared with the previous study applying a 4-second window \cite{dedeyne_exploring_2021}, the f1-scores in this study were higher than 0.95 for both datasets. With a small window size, each activity was segmented individually. However, most individual OEP activity was almost the same as ADLs. For example, \textit{forwards walking} in the OEP was the same as walking in daily life.\par

A large window size, on the other hand, included multiple exercises instead of a single one. Therefore, one segment contained more information. As shown in Fig.~\ref{fig:signal}, each OEP sub-class took a shorter duration than ADLs. Also, the signals of OEP showed higher variance than ADLs. Therefore, a large sliding window could better identify these characteristics. A 10-minute sliding window with 75\% overlap resulted in a delay of 2.5 minutes, which was applicable for offline systems. In real life, the system could also be applied as a reminder to perform OEP for older adults. \par

The annotation of OEP was from the beginning of the first OEP activity to the end of the last OEP activity. During the program, the subjects could take a rest on a chair or go to the bathroom. These behaviors are also expected in real life. Therefore, the smoothing algorithm was applied to also take non-OEP activities into account and hence, the recognition performance was improved.\par

In stage 1, CNN-BiLSTM was proven to be the best classification model for both datasets. The scenarios using a large window (long time series) for OEP recognition could take advantage of CNN-BiLSTM. Similar results were also reported by a previous study, reporting that CNN-LSTM outperformed CNN, LSTM, and Transformer for IMU-based activity recognition \cite{trujillo2023accuracy}. However, deep learning models require more training examples than classical models. Thus, the models still encountered overfitting problems due to dataset size. Although the usage of dropout layers improved this problem and made the models more efficient than hand-crafted features, the lack of training examples was still a limitation of the study.
\par

The amount of ADLs collection was limited. In dataset 1, the ADLs included \textit{walking}, \textit{standing}, \textit{sitting}, and \textit{cycling}. In dataset 2, the duration of ADLs was no more than two hours. In real life, ADLs happen much more than OEP. Therefore, in the future, the methods should be validated on the dataset with more types and longer duration of ADLs. Another limitation was that the study did not consider calibration and misplacement of the sensor, since the older adults could not comply with these tasks alone. Although it would result in decreased f1-scores, the system still showed robustness in generalization according to Fig.~\ref{fig:box}.

\subsection{Stage 2}

The size of the training set was small compared with the number of OEP classes. To reduce the number of classes for each model, a hierarchical classification method was applied. Considering the transition activities between each two OEP sub-classes, the f1-scores of some sub-classes were decreased such as \textit{sideways walking}, \textit{walking stairs}, \textit{walking turn}, and \textit{sit-to-stand}. These activities could also happen in ADLs. Therefore, some transition activities were classified as the ones in OEP sub-classes, although such influence was improved by post-processing in stage 2. Since the results of stage 2 were based on the output of stage 1, they showed higher variance according to Fig.~\ref{fig:box}. \par

The f1-scores of \textit{knee bends} and \textit{ankle plantarflexors} were increased in dataset 2. The first reason was that dataset 1 was also applied for training to test dataset 2. With more training data, the models were less over-fitting. The second reason was that the subjects in dataset 2 were younger than in dataset 1, as shown in Table~\ref{tab:subinfo}. Therefore, the subjects in dataset 2 performed the exercises with less intra-class variance.\par

On the other hand, compared to dataset 1, exercises such as \textit{abdominal muscles} and \textit{sit to stand} had decreased f1-scores in dataset 2. Such a decrease was due to the fact that the subjects followed the booklet rather than the direct instructions from the therapists. Therefore, they were unable to adhere to the instructions effectively.\par

Because of the lack of training data, deep learning models were not applied in stage 2, which limited this study. In future studies, more data will be collected so that deep learning models can be applied to improve performance.

\subsection{Comparison with previous studies}
There were only two studies applying wearable sensors for OEP recognition. The first study applied a waist-mounted IMU to recognize OEP and ADLs for (sarcopenic) older adults \cite{dedeyne_exploring_2021}. The study categorized OEP exercises into strength, balance, and other exercises, without classifying the OEP sub-classes. The f1-scores for classifying OEP and ADLs of the study are shown in Table~\ref{tab:compare1}. The results show that our proposed 10-minute sliding window and CNN-BiLSTM model obtained higher f1-scores than the previous study using a 4-second sliding window and a RF model. Besides the advantages of the deep learning models, the large sliding window also shows its ability to capture more information, leading to improved results.

\begin{table}[!t]
\caption{Comparison: OEP f1-scores of the previous study and this study}
\label{tab:compare1}
\center
\begin{tabular}{|l|l|l|}
\hline
previous study \cite{dedeyne_exploring_2021} & Dataset 1 & Dataset 2 \\ \hline
0.777                                        & 0.983     & 0.950     \\ \hline
\end{tabular}
\end{table}

The second study applied five IMUs on the four limbs and waist worn by young adults \cite{bevilacqua_human_2019}. Since the study categorized OEP exercises differently, only four classes were comparable to this study. Also, although the study applied five IMUs, only the results with the waist-mounted IMU were compared to this study. Table~\ref{tab:compare2} compares the f1-scores of the previous study and this study. The results showed that the previous study (only validated on four healthy young subjects) did not obtain f1-scores over the threshold (0.8, which is a common threshold value for OEP \cite{dedeyne_exploring_2021}) for the four classes. On the other hand, our proposed method obtained higher f1-scores for these classes, with two of them (\textit{knee bends} and \textit{sit to stand}) exceeding the threshold. Additionally, the previous study did not take into account the transition activities between OEP sub-classes. The superior performance of our proposed method may be attributed to the two-level labeling of OEP sub-classes and relevant hand-crafted features such as relative start time.

\begin{table}[!t]
\caption{Comparison: f1-scores for OEP sub-classes of the previous study (with the waist-mounted IMU) and this study}
\label{tab:compare2}
\center
\begin{tabular}{|l|l|l|l|l|}
\hline
                                                                 & ankle dor & knee bends & sideways & sit to stand \\ \hline
previous study \cite{bevilacqua_human_2019}                       & 0.370     & 0.310      & 0.790    & 0.490        \\ \hline
\begin{tabular}[c]{@{}l@{}}This study: \\ dataset 1\end{tabular} & 0.478     & 0.722      & 0.565    & \textbf{0.869}        \\ \hline
\begin{tabular}[c]{@{}l@{}}This study: \\ dataset 2\end{tabular} & \textbf{0.764}     & \textbf{0.836}      & \textbf{0.744}    & 0.840        \\ \hline
\end{tabular}
\end{table}

\section{Conclusion}
\label{sec:conclusion}

This study proposes a hierarchical system to recognize OEP from a waist-mounted IMU. The system was tested on the older adults in both lab and home environments. In stage 1, using the CNN-BiLSTM architecture, the system could distinguish OEP and ADLs with f1-scores over 0.95. The results showed the capability of a single IMU to evaluate the compliance of OEP for older adults with and without sarcopenia. Besides, the system could distinguish four OEP sub-classes with f1-scores over 0.8: \textit{ankle plantarflexors}, \textit{abdominal muscles}, \textit{knee bends}, and \textit{sit-to-stand}. The results showed the potential of monitoring the compliance of OEP using a single IMU in daily life. Furthermore, the proposed system demonstrated its capability to recognize and analyze OEP sub-classes. \par

In the future, the size of the dataset should be improved, since the amount of training examples did not support deep learning models in stage 2. Also, the hyperparameters of this system could be reduced for better generalization. For example, end-to-end deep learning models such as temporal convolutional networks (TCN) \cite{farha_ms-tcn_2019} could be applied without post-processing. 

\section*{Acknowledgment}

This work was supported by the China Scholarship Council (CSC). The ENHANce project (S60763) received a junior research project grant from the Research Foundation Flanders (FWO) (G099721N). The funding provider did not contribute or influence the design of the study and data collection, analysis and interpretation in writing this manuscript.

\section*{REFERENCES}

\bibliographystyle{IEEEtran}
\bibliography{main}

\end{document}